\documentclass[
]{ceurart}

\usepackage{listings}
\usepackage[ruled,vlined,linesnumbered]{algorithm2e}
\usepackage[frozencache=true,cachedir=_minted-paper]{minted}
\usepackage{caption}
\usepackage[section]{placeins}
\usepackage{wrapfig}
\usepackage{subcaption}

\begin{document}

\copyrightyear{2023}
\copyrightclause{Copyright for this paper by its authors.
  Use permitted under Creative Commons License Attribution 4.0
  International (CC BY 4.0).}

\conference{In A. Martin, K. Hinkelmann, H.-G. Fill, A. Gerber, D. Lenat, R. Stolle, F. van Harmelen (Eds.), 
Proceedings of the AAAI 2023 Spring Symposium on Challenges Requiring the Combination of Machine Learning and Knowledge Engineering (AAAI-MAKE 2023), Hyatt Regency, San Francisco Airport, California, USA, March 27-29, 2023.}

\title{Neuro-symbolic Rule Learning in Real-world Classification Tasks}

\author[1]{Kexin Gu Baugh}[%
email=kexin.gu17@imperial.ac.uk,
orcid=0000-0003-3680-8442,
]

\author[1]{Nuri Cingillioglu}[%
email=nuric@imperial.ac.uk,
]

\author[1]{Alessandra Russo}[%
email=a.russo@imperial.ac.uk,
]

\address[1]{Imperial College London}

\begin{abstract}
Neuro-symbolic rule learning has attracted lots of attention as it offers better interpretability than pure neural models and scales better than symbolic rule learning. A recent approach named pix2rule proposes a neural Disjunctive Normal Form (neural DNF) module to learn symbolic rules with feed-forward layers. Although proved to be effective in synthetic binary classification, pix2rule has not been applied to more challenging tasks such as multi-label and multi-class classifications over real-world data. In this paper, we address this limitation by extending the neural DNF module to (i) support rule learning in real-world multi-class and multi-label classification tasks, (ii) enforce the symbolic property of mutual exclusivity (i.e. predicting exactly one class) in multi-class classification, and (iii) explore its scalability over large inputs and outputs. We train a \textit{vanilla neural DNF model} similar to pix2rule's neural DNF module for multi-label classification, and we propose a novel extended model called \textit{neural DNF-EO} (Exactly One) which enforces mutual exclusivity in multi-class classification. We evaluate the classification performance, scalability and interpretability of our neural DNF-based models, and compare them against pure neural models and a state-of-the-art symbolic rule learner named FastLAS. We demonstrate that our neural DNF-based models perform similarly to neural networks, but provide better interpretability by enabling the extraction of logical rules. Our models also scale well when the rule search space grows in size, in contrast to FastLAS, which fails to learn in multi-class classification tasks with 200 classes and in all multi-label settings. \end{abstract}


\maketitle

\section{Introduction}

The recent success of deep learning has started to bring transformative impacts in different sectors of our society. However, they are not interpretable and cannot be used to solve real-world problems that require safe and sound reasoning. There has been an increasing interest in neuro-symbolic approaches that combine symbolic reasoning with neural networks \cite{nsai-3-wave}. Some of these approaches have proposed differentiable models that support inductive learning of symbolic rules~\cite{pix2rule, delta-ilp, lri, lnn-ilp, hri, nlm, dnl-ilp, dlm}. Similarly to pure logic-based learning (e.g. Inductive Logic Programming (ILP) \cite{ilp}), these differentiable ILP methods aim to learn interpretable rules from positive and negative examples, but they do so using pure statistical mechanisms. Most approaches have been shown to perform well over classical ILP benchmark tasks where data are symbolic~\cite{delta-ilp, lri, hri, nlm, dnl-ilp, dlm}, whereas a few have been applied to unstructured data for learning rules capable of solving classification tasks~\cite{pix2rule}. Many of the proposed differentiable ILP approaches rely on human-engineered rule templates to restrict their search space~\cite{delta-ilp, lri, lnn-ilp, hri}, and these templates are often limited in their expressivity since they allow predominantly Datalog chain rules. Approaches that do not require rule templates either lose interpretability~\cite{nlm}, or resort to additional tricks, e.g., to treat negation in a specific way~\cite{dlm}. On the other hand, \textit{pix2rule} \cite{pix2rule} presents a neural model capable of learning \textit{interpretable} rules from unstructured data in an end-to-end fashion \textit{without} the use of template rules. The model uses a \emph{Disjunctive Normal Form (DNF)} module that has two feed-forward layers and enables learning over a large search space of interdependent rules with negation. Although shown to be effective in solving binary classification tasks, it has only been empirically evaluated over synthetic data.

In this paper, we build upon pix2rule's neural DNF module to support rule learning in \textit{real-world multi-class} and \textit{multi-label} classification tasks. We investigate the learning capability of the \emph{vanilla neural DNF model} similar to the neural DNF module in pix2rule on real-world multi-label classification tasks. We also propose a novel model called \emph{neural DNF-EO (Exactly-One)} for multi-class classification to address its requirement of mutual exclusivity, i.e. having exactly one class to be true at a time. In a symbolic learner, logical constraints enforce mutual exclusion, but a vanilla MLP is not able to strictly enforce it with a cross-entropy loss. Our neural DNF-EO model is trained with a cross-entropy loss but enforces mutual exclusion, realising the symbolic constraint in a differentiable fashion. We evaluate our models over two real-world datasets, the multi-class dataset CUB-200-2011 \cite{cub-200-201} and its subsets, and the multi-label dataset TMC2007-500 \cite{tmc} and its subsets. We focus on exploring (i) the classification performance and scalability of our models over an increasing number of discrete attributes, and (ii) the interpretability of the learned rules versus other pure symbolic and statistical methods such as decision trees. Our experimental results show that our neural DNF-based models perform as well as pure neural networks, whilst providing symbolic rules through post-training processes. They scale better than a state-of-the-art pure symbolic rule learner FastLAS \cite{fastlas} over large search spaces. Our models successfully solve all tasks while FastLAS fails to learn in multi-class classification tasks with 200 classes and in all multi-label settings. Compared to other differentiable ILP methods (e.g.,\cite{delta-ilp, nlm, dlm}), our neural DNF-based models do not require human-engineered rule templates, provide interpretability by enabling extraction of symbolic rules after training, and handle logical negation without special treatment.

\section{Background} \label{chap:background}

This chapter provides a summary of the neural Disjunctive Normal Form (neural DNF) module presented in pix2rule \cite{pix2rule}. pix2rule proposes a novel semi-symbolic layer that behaves like a neuro-symbolic disjunction or conjunction of (negated) atoms. Its disjunctive or conjunctive behaviour is controlled by a bias parameter. Given some continuous input $x_i \in [\bot, \top]$ where $\top = 1$ and $\bot = -1$, the semi-symbolic layer is defined as:

\begin{align}
    y &= f(\sum_{i} w_i x_i + \beta) \label{eq:ss-y}\\
    \beta &= \delta (\max_{i} |w_i| - \sum_i |w_i|) \label{eq:ss-bias}
\end{align}

\noindent where $w_i$ are trainable weights, $f$ is the activation function $\tanh$, and $y$ is the layer output. By setting $\delta$ to be 1 or -1, the layer can approximate conjunction or disjunction. Using this formulation instead of a t-norm fuzzy logic with a continuous input space of [0, 1] attempts to reduce the gradient vanishing problem by avoiding the multiplication of many close-to-zero values \cite{pix2rule}. The neural DNF module is created by stacking a conjunctive semi-symbolic layer followed by a disjunctive semi-symbolic layer. During training, the magnitude (i.e. absolute value) of the $\delta$ parameters in both layers, which controls the magnitude of logical biases on the layers, is increased gradually from 0.1 to 1, so that the module can learn more stably and slowly adjust to the logical biases. Symbolic rules can be extracted from a trained neural DNF module through post-training processes. The rules extracted are expressed in a logic programming language called Answer Set programming (ASP)~\cite{asp}. Although shown to be effective in learning end-to-end ASP rules from unstructured synthetic binary datasets, it is unclear whether the neural DNF module would still learn well in real-world multi-class and multi-label classification tasks.

\section{Neural DNF-based Models}

Based on the semi-symbolic layers and the neural DNF module proposed in pix2rule \cite{pix2rule}, we train two types of neural DNF-based models: a \textit{vanilla neural DNF model} for multi-label classification, and a novel \textit{neural DNF-EO model} for multi-class classification.

\vspace{0.5em}

\captionsetup{font=normalfont, labelfont=bf}

\begin{wrapfigure}{r}{.45\textwidth}
\vspace{-15pt}
\includegraphics[width=\linewidth]{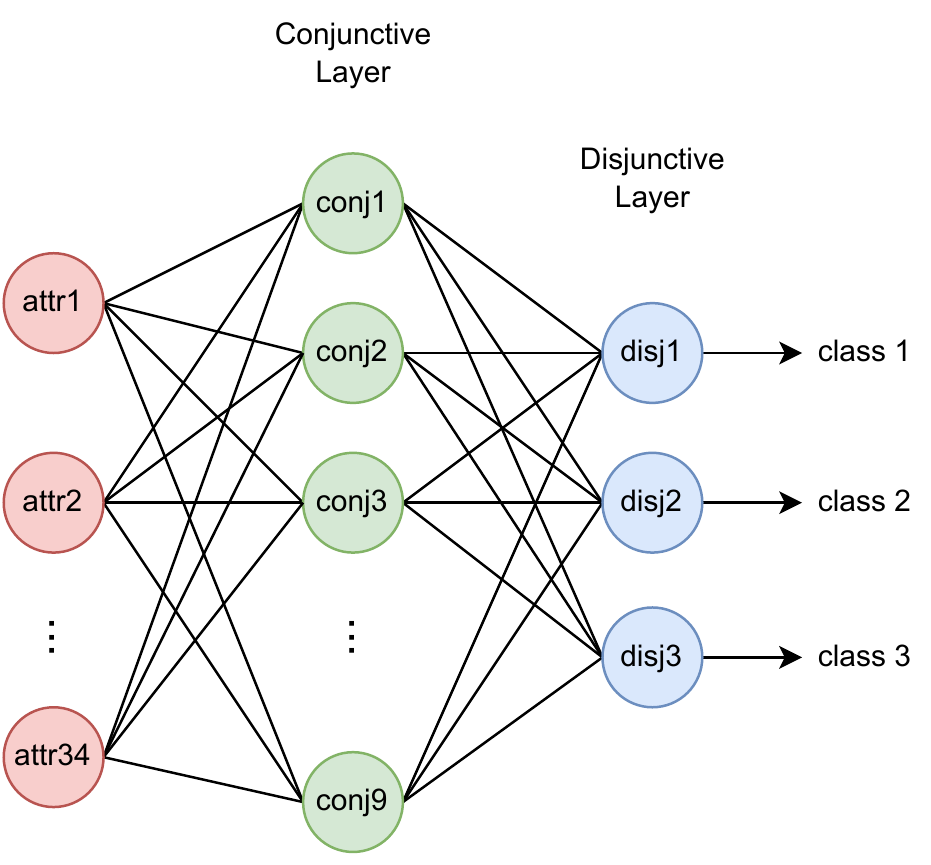}
\captionof{figure}{An example of a vanilla neural DNF model.}
\vspace{-18pt}
\label{fig:vanilla-dnf}
\end{wrapfigure}

The \textbf{vanilla neural DNF model} as shown in Figure~\ref{fig:vanilla-dnf} consists of a semi-symbolic conjunctive layer followed by a disjunctive one, sharing the same architectural design as pix2rule's neural DNF module. It is suitable for independent rule learning in both binary and multi-label classification tasks. In the latter case, the target (binary) predicates are independent of each other.

\vspace{0.5em}

\noindent \textbf{The challenge of multi-class classification} \label{chap:cross-entropy-and-multi-class-classification}

A multi-class classification task requires exactly one class to be true at a time. This can be written as a logical constraint expressed as follows: \mbox{$\leftarrow class(X), class(Y), X\neq Y$}. It states that two different classes cannot be true at the same time. This constraint is not directly satisfiable by an MLP trained with a cross-entropy loss. An MLP trained under a cross-entropy loss takes the $\arg \max$ of the output when predicting a class at inference time and does not need to enforce mutual exclusion over its outputs. Hence, it is not capable of capturing the logical constraint. Similarly, training our vanilla neural DNF model with a cross-entropy loss does not enforce mutual exclusivity on the model. At inference time, a neural DNF-based model's $\tanh$ output of a class is treated as True ($\top$) if the value is greater than 0 and otherwise False ($\bot$). Since the softmax function in the cross-entropy loss only amplifies the difference between the output logits of each class, it does not have the concept of a value being True if it goes above a threshold, which is required for us to interpret our neural DNF-based model's outputs. The vanilla neural DNF model's output logits under the cross-entropy loss are only trained to be different from each other and not to be extreme values that fall on different sides of our desired threshold value. For example, if the raw output logits of our vanilla neural DNF model are $\tilde{y} = [-2, 1.5, 3]$ with a target class of 2 (zero-indexed), then $\text{softmax}(\tilde{y}) = [0.006, 0.181, 0.813]$ which is very different between each pair, and the cross-entropy loss would be $0.207$. However, the $\tanh$ output would be $\tanh(\tilde{y}) = [-0.964, 0.905, 0.995]$, which would be interpreted as $[\bot, \top, \top]$ and violate the `exactly-one-class' constraint in the multi-class setting. Our experiments of using cross-entropy loss to train vanilla neural DNF models on multi-class classification tasks (in Appendix~\ref{app:vndnf-cub}) show this issue of multiple classes being predicted as true, by interpreting the output to be discrete True and False value at threshold value 0. This motivates our second model, which (i) can be trained with cross-entropy loss, even though it only pushes output logits to be different; (ii) but still have the property of mutual exclusivity guaranteed.

\vspace{0.5em}

\begin{wrapfigure}{r}{.55\textwidth}
\vspace{-15pt}
\includegraphics[width=\linewidth]{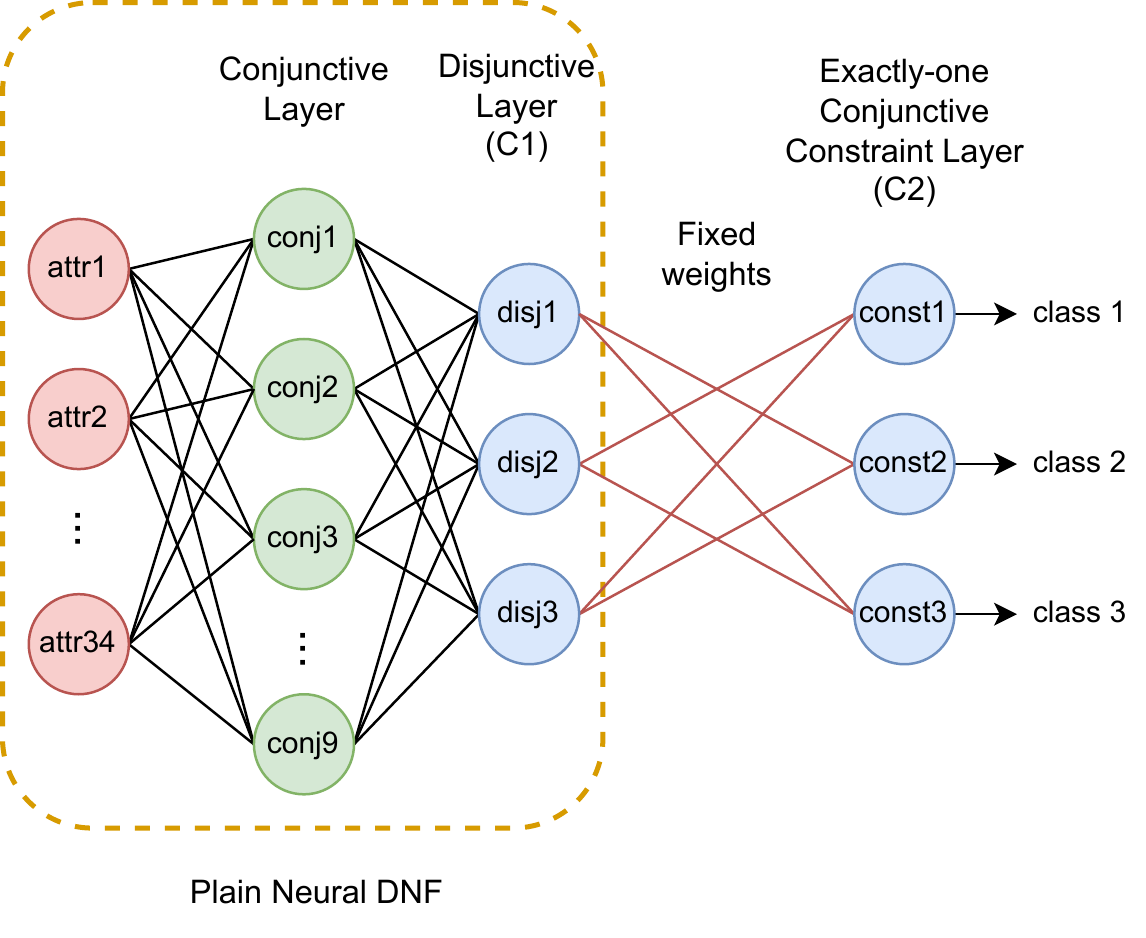}
\captionof{figure}{An example of a neural DNF-EO model. The plain neural DNF in the dotted yellow box is identical to a vanilla neural DNF model.}
\label{fig:dnf-eo}   
\vspace{-18pt}
\end{wrapfigure}

The \textbf{neural DNF-EO model} shown in Figure~\ref{fig:dnf-eo} has a constraint layer (layer C2 in the figure) that reflects the `exactly-one' constraint `$class_i \leftarrow \land_{j, j \neq i} \text{ not } class_j$' which enforces mutual exclusion on the plain neural DNF (in the dotted yellow box in the figure). If more than one output is true after the disjunctive layer C1, no class would be true after the constraint layer C2. This constraint layer C2 is initialised with all connections being -6\footnote{Having weights to be 6 or -6 saturates $\tanh$ ($\tanh(\pm 6)\approx \pm1$), and the output should be approximately 1 or -1.}, representing the logical negation `not', and its weights are not updated. During training, the constraint layer C2's output logits are pushed to be different and to match with the expected class by the cross-entropy loss, but the output of disjunctive layer C1 of the plain neural DNF is pushed to have only one value being true and all the rest being false. At inference time, we remove the constraint layer C2 and inference with the plain neural DNF that is enforced to output exactly one class true at a time, thus providing mutual exclusivity. 

\vspace{0.5em}

In order to extract rules from these neural DNF-based models, we follow and expand on the post-training processes used in pix2rule. Our post-training pipeline consists of 4 stages: pruning, finetuning, thresholding and extraction. Pruning, thresholding and extraction remain mostly the same as in pix2rule except for the metrics we use, while we add finetuning as a new step to adjust for learning and extracting rules in real-world datasets.

\textbf{Pruning} process sets the weights that do not heavily affect the performance of the neural DNF-based model to 0, where the accepted performance drop is controlled by a hyperparameter $\epsilon$. We prune the disjunctive layer first followed by the conjunctive layer. We also remove any disjunct that uses a conjunction with an empty body, i.e. the conjunction has weight 0 for all its conjuncts. For the neural DNF-EO model, we remove the constraint layer C2 and prune only the plain neural DNF.

\textbf{Finetuning} process adjusts the model's weights by re-training the neural DNF-based model on the training set without updating the pruned weights. For the neural DNF-EO model, we add back the constraint layer after the pruned plain neural DNF during the finetuning process to maintain mutual exclusivity.

\textbf{Thresholding} process goes through a range of possible threshold values, such that if the magnitude of the weight is greater than that threshold, the magnitude of the weight would be set to 6. By setting the value of weights to 6 or -6, $\tanh$ output can be saturated close to 1 and -1. We choose the threshold value that gives the best performance based on the validation set. Again, the constraint layer C2 in the neural DNF-EO model is dropped and the thresholding is done on the plain neural DNF.
    
\textbf{Symbolic rule extraction} is done by taking input attributes/conjunctions connected to weights with value $\pm 6$ as the conjunct/disjunct of their connected conjunction/disjunction. And each disjunct of the disjunction can be treated as a separate rule to classify a class or label. We express the extracted rules as an ASP program.

\section{Datasets}
\label{sec:dataset}

We experiment on two datasets: CUB-200-2011 \cite{cub-200-201} and TMC-2007-500 \cite{tmc}, to explore multi-class and multi-label classification respectively. Both provide discrete attributes that are translated to propositional atoms and taken as input for our neural DNF-based models.

\subsection{CUB Dataset}

\begin{figure}[h]
    \centering
    \includegraphics[width=0.8\textwidth]{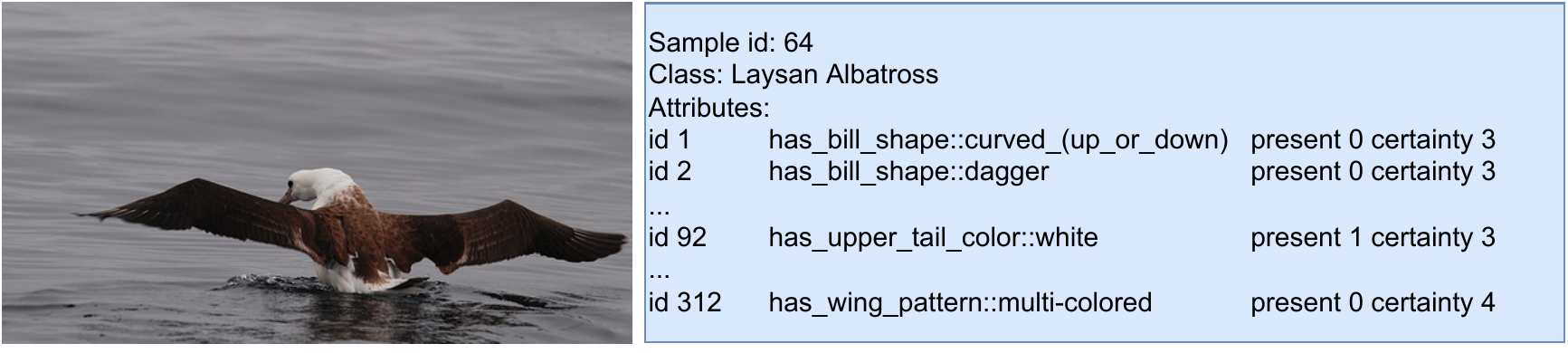}
    \caption{A sample from the CUB-200-2011 dataset before any data pre-processing. We ignore the image and the input to our neural DNF-based models would only be the attributes.}
    \label{fig:cub-dataset-demo}
\end{figure}

CUB-200-2011 \cite{cub-200-201} is a real-world multi-class bird classification dataset with crowd-worker-annotated bird attributes as concepts. A sample such as in Figure~\ref{fig:cub-dataset-demo} contains an image, a set of 312 attribute annotations and a class label out of 200 possible classes. To focus on scalable and interpretable rule learning form propositions, we leave the end-to-end learning from images as future work.

The attribute annotation has two components. The `present' label represents whether the sample has that attribute, where 0 means attribute not present and 1 means attribute present. The other one is the certainty label which indicates the degree of confidence of the `present' label being true. The certainty goes from 1 to 4 where 1 is not visible, 2 is guessing, 3 is probably and 4 is definitely. These attribute annotations are not consistent throughout the dataset. To de-noise it, we follow the same pre-processing procedure used in the Concept Bottleneck Models paper \cite{cbm}: compute the median of an attribute's presence label in a class, and keep it if this attribute is consistently present in at least $N$ classes, where $N$ is a hyperparameter. With different values of $N$, the data pre-processing procedure results in different numbers of attributes remaining. This procedure removes the certainty labels and unifies the attribute encoding for different samples in the same class. We provide a pseudo-code for the pre-processing procedure and further discussion in Appendix~\ref{app:data-pre-processing-cub}.

Table~\ref{tab:cub-different-dataset} provides the statistics of the CUB dataset (before and after pre-processing) and the pre-processed subsets we experiment on. These data subsets are created for the purpose of scalability check, and are generated by taking samples of the first $k$ classes (class 1 to $k$) in the original dataset, e.g. class 1 to 3 for CUB-3 subset. We apply the data pre-processing procedure with different values $N$ due to the different numbers of classes in the subsets. All our multi-class classification experiments use pre-processed CUB datasets/subsets, and, for simplicity, from now on we do not explicitly mention `after pre-processing' for each CUB dataset/subset.

\begin{table}[h]
\begin{minipage}[b]{.5\linewidth}
\tiny
\begin{tabular}{c|ccc}
Dataset/Subset                      & No. samples & No. classes & No. attributes \\ \hline
Original CUB-200-2011                  & 11788             & 200               & 312                  \\
CUB-3    & 178               & 3                 & 34                   \\
CUB-10   & 543               & 10                & 41                   \\ 
CUB-15   & 837               & 15                & 40                   \\ 
CUB-20   & 1115              & 20                & 48                   \\ 
CUB-25   & 1402              & 25                & 50                   \\ 
CUB-50   & 2889              & 50                & 61                   \\ 
CUB-100  & 5864              & 100               & 82                   \\
CUB-200  & 11788             & 200               & 112                  \\
\end{tabular}
\subcaption{Statistics of different CUB datasets/subsets.}
\label{tab:cub-different-dataset}
\end{minipage}%
\begin{minipage}[b]{.5\linewidth}
\tiny
\begin{tabular}{c|ccc}
Dataset/Subset           & No. samples    & No. labels & No. attributes \\ \hline
Original TMC2007-500     & 28596          & 22             & 500\\
TMC-3  & 16947           & 3              & 59                 \\
TMC-5  & 18565           & 5              & 60                 \\
TMC-10 & 21454           & 10             & 34                 \\
TMC-15 & 24318           & 15             & 80                 \\
TMC-22 & 28538           & 22             & 103                \\
\end{tabular}
\subcaption{Statistics of different TMC datasets/subsets.}
\label{tab:tmc-different-datasets}
\end{minipage}
\caption{Data statistics for CUB and TMC dataset/subsets. Except for `Original CUB-200-2011' and `Original TMC2007-500', all the other datasets/subsets are pre-processed. The number of samples is the sum of samples from all train, validation and test sets. We use the suffix `-k' after the dataset name to represent the number of classes/labels used in the datasets/subsets.}
\end{table}

\subsection{TMC2007-500}

\begin{figure}[h]
    \centering
    \includegraphics[width=0.7\linewidth]{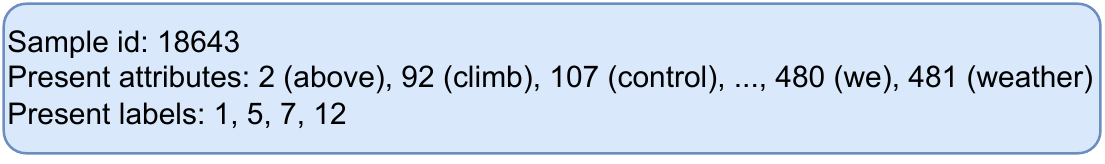}
    \caption{A sample from the TMC2007-500 dataset before any data pre-processing. The dataset does not explicitly give the names of the anomalies for each label.}
    \label{fig:tmc-sample}
\end{figure}

TMC2007-500 \cite{tmc} is a texted-based multi-label classification dataset based on space shuttle reports, and the goal is to identify different types of anomalies from the text. The dataset provides 500 discrete attributes along with 22 labels, while the reports and actual anomalies are not revealed due to their sensitivity. A sample from the dataset is shown in Figure~\ref{fig:tmc-sample}. The attributes themselves are words, making the input similar to an NLP bag-of-words encoding where each bit of the encoding represents if a word appears in the report. For this reason, we apply data pre-processing (detailed in Appendix~\ref{app:data-pre-processing-tmc}) to filter out attributes with low mutual information (MI) with respect to the output label combinations (not on individual labels). Only attributes with MI values greater or equal to a threshold $t$ are kept.

Similar to the CUB dataset, we create subsets and apply the pre-processing on each subset. Table \ref{tab:tmc-different-datasets} shows the statistics of the datasets/subsets after pre-processing. We apply different MI thresholds during pre-processing on different subsets, resulting in different numbers of attributes used in each subset.

\section{Experiment Results}

We evaluate our neural DNF-based models in terms of classification performance, scalability, and interpretability. The classification task can be treated as a propositional ILP task with no background knowledge. We include a two-layer MLP as a baseline for comparison to a pure neural method, and a state-of-the-art symbolic learner FastLAS \cite{fastlas} for direct comparison between symbolic and neuro-symbolic approaches. We also compare our models against a decision tree since it is a simple classifier that also provides interpretability.

\subsection{CUB dataset/subsets}

For multi-class classification tasks, we use the Jaccard index score to measure the correct prediction of a single class at a time in case our neural DNF-EO model outputs multiple classes for one sample. If the model predicts exactly one class, the Jaccard score is of value 1; but if two or more classes are predicted, the Jaccard score will be less than 1. In each CUB experiment, we train the neural DNF-EO model with a cross-entropy loss function and let the constraint layer enforce mutual exclusivity. At inference time, we remove the constraint layer (layer C2 in Figure~\ref{fig:dnf-eo}) from the neural DNF-EO model and keep only the plain neural DNF. The plain neural DNF outputs a value in the $[-1, 1]$ range for each class, and we take the class as True if the output value is greater than 0. Table~\ref{tab:dnf-eo-cub} shows the performance of the neural DNF-EO model after training and at different stages of the post-training pipeline. The plain neural DNF performs correctly with mutual exclusion enforced on all test sets, as shown in the `\textit{Train Jacc. (no C2) column}' in Table~\ref{tab:dnf-eo-cub}. It also provides ASP rules like in Table~\ref{tab:dnf-eo-rules-cub-3} after the post-training pipeline. However, after the CUB-20 subset, the performance difference between a thresholded plain neural DNF (`\textit{Threshold Jacc. (no C2)}' column) and the original after training (`\textit{Train Jacc. (no C2)}' column) increases along with the number of classes in the subset. Upon close inspection, we believe the issue is about maintaining mutual exclusivity with an increasing number of classes using the current thresholding process.

\begin{table}[h]
\begin{tabular}{c|cccccc}
Dataset & \begin{tabular}[c]{@{}c@{}}Train\\ Acc.\end{tabular} & \begin{tabular}[c]{@{}c@{}}Train Jacc.\\ (no C2)\end{tabular} & \begin{tabular}[c]{@{}c@{}}Prune Jacc.\\ (no C2)\end{tabular} & \begin{tabular}[c]{@{}c@{}}Finetune\\ Jacc.\\ (no C2)\end{tabular} & \begin{tabular}[c]{@{}c@{}}Threshold\\ Jacc.\\ (no C2)\end{tabular} & \begin{tabular}[c]{@{}c@{}}Rules\\ Jacc.\end{tabular} \\ \hline
CUB-3   & 1.000                                                & 1.000                                                         & 1.000                                                         & 1.000                                                              & 1.000                                                               & 1.000                                                 \\
CUB-10  & 1.000                                                & 1.000                                                         & 1.000                                                         & 1.000                                                              & 1.000                                                               & 1.000                                                 \\
CUB-15  & 1.000                                                & 1.000                                                         & 1.000                                                         & 1.000                                                              & 1.000                                                               & 1.000                                                 \\
CUB-20  & 1.000                                                & 1.000                                                         & 1.000                                                         & 1.000                                                              & 1.000                                                               & 1.000                                                 \\
CUB-25  & 1.000                                                & 1.000                                                         & 1.000                                                         & 1.000                                                              & 0.935                                                               & 0.933                                                 \\
CUB-50  & 1.000                                                & 1.000                                                         & 0.982                                                         & 0.975                                                              & 0.641                                                               & 0.636                                                 \\
CUB-100 & 1.000                                                & 1.000                                                         & 0.967                                                         & 0.978                                                              & 0.630                                                               & 0.632                                                 \\
CUB-200 & 1.000                                                &  1.000                                                        & 0.967                                                         & 0.948                                                               & 0.225                                                              & 0.225
\end{tabular}
\caption{\textbf{Neural DNF-EO model} trained with \textbf{cross-entropy loss} in different CUB subsets/dataset, at different stages of training and post-training pipeline. All metrics are calculated by averaging across the test set. (`Acc.' short for accuracy, `Jacc.' short for Jaccard index score)}
\label{tab:dnf-eo-cub}
\end{table}

\begin{table}[h]
\begin{minipage}{.4\linewidth}
\begin{minted}{text}
black_footed_albatross :-
  not has_bill_colour_black,
  not has_wing_pattern_solid.
laysan_albatross :-
  has_crown_colour_white.
sooty_albatross :-
  not has_bill_colour_buff,
  has_crown_colour_black,
  not has_crown_colour_white.
\end{minted}
\end{minipage}%
\hspace{0.05\linewidth}%
\begin{minipage}{0.55\linewidth}
\raggedleft
\begin{tabular}{c|ccc}
Dataset & MLP   & \begin{tabular}[c]{@{}c@{}}Decision Tree\\ (no pruning)\end{tabular} & \begin{tabular}[c]{@{}c@{}}Neural DNF-EO\\ without C2\\ (after training)\end{tabular} \\ \hline
CUB-3   & 1.000 & 1.000                                                                & 1.000                                                                               \\
CUB-10  & 1.000 & 1.000                                                                & 1.000                                                                               \\
CUB-15  & 1.000 & 1.000                                                                & 1.000                                                                               \\
CUB-20  & 1.000 & 1.000                                                                & 1.000                                                                               \\
CUB-25  & 1.000 & 1.000                                                                & 1.000                                                                               \\
CUB-50  & 1.000 & 1.000                                                                & 1.000                                                                               \\
CUB-100 & 1.000 & 1.000                                                                & 1.000                                                                               \\
CUB-200 & 1.000 & 1.000                                                                & 1.000                                                                              
\end{tabular}
\end{minipage}
\par
\begin{minipage}[t]{.4\linewidth}
\caption{ASP rules extracted from a trained neural\\DNF-EO model on CUB-3 experiment.}
\label{tab:dnf-eo-rules-cub-3}
\end{minipage}%
\hspace{0.05\linewidth}%
\begin{minipage}[t]{0.55\linewidth}
\caption{Average test accuracy comparison among MLP, decision\\tree (no pruning) and trained neural DNF-EO model (no\\constraint layer C2, no post-training processing), across\\different CUB subsets/dataset.}
\label{tab:cub-acc-comp}
\end{minipage}
\end{table}

\noindent \textbf{Comparison to baselines}\ \ \ \ For the baseline methods, we train the two-layer MLP identically as our neural DNF-EO model but without the $\delta$ parameter adjustment. We specify FastLAS's scoring function to be on body length so that FastLAS learns the most compact rules.

\textbf{Accuracy comparison}\ \ \ \ We measure the accuracy of an MLP and a trained plain neural DNF (without the constraint layer C2) by taking the $\arg \max$. A trained decision tree gives one class as the prediction after following the decision path and reaching the leaf node. All three models perform equally perfectly across all experiments, as shown in Table~\ref{tab:cub-acc-comp}.

\textbf{Jaccard index score and scalability comparison}\ \ \ \ We interpret the $\tanh$ output of the plain neural DNF to be True if the value is greater than 0, giving the possibility of multiple class being True at the same time. Both the ASP rules learned by FastLAS and the ASP rules extracted from the plain neural DNF after the post-training pipeline can give more than 1 class in an answer set. Thus we choose the Jaccard index score as a measurement of mutual exclusivity for comparing these three approaches. We also attempt to interpret the 2-layer MLP to see if there is any mutual exclusivity at all: we expect some similarity between the MLP and our neural DNF-EO model as both models have $\tanh$ as the activation function at hidden layers and their weights are initialised the same. Thus our attempt at interpreting the MLP follows the same way as we interpret our neural DNF-EO model by thresholding at 0 after $\tanh$. However, as discussed in Chapter~\ref{chap:cross-entropy-and-multi-class-classification}, softmax only pushes the MLP to have different output logits and no need to push the logits towards a threshold such that they have symbolic meaning. Thus we expect the MLP to have poor Jaccard index scores. Table~\ref{tab:cub-jacc-comp} shows the average test Jaccard score of all three (neuro-)symbolic approaches as well as our symbolic interpretation attempt on MLP. As expected, the MLP displays little to no symbolic mutual exclusivity for tasks with 10 classes and onwards. FastLAS learns perfect mutually-exclusive rules for all subsets but fails to learn from the full dataset of CUB-200, where it runs out of memory when computing the opt-sufficient search space. Our trained plain neural DNF without any post-training processing learns perfectly in all settings regardless of the number of classes. This shows that our model provides mutual exclusion enforcement just like FastLAS, but also better scalability as it does not suffer from hypothesis search space growth. However, we cannot always extract well-performed rules from it, as there is a performance gap between the rules and the plain neural DNF without constraint layer C2 from CUB-25 and onwards.

\begin{table}[h]
\begin{tabular}{c|ccc|c}
Dataset & FastLAS & \begin{tabular}[c]{@{}c@{}}Neural DNF-EO without C2\\ (after training)\end{tabular} & \begin{tabular}[c]{@{}c@{}}Rules Extracted from\\ Neural DNF-EO without C2\end{tabular} & \begin{tabular}[c]{@{}c@{}}MLP\\ (0-thresholded)\end{tabular} \\ \hline
CUB-3   & 1.000   & 1.000                                                                               & 1.000                                                                                   & 1.000                                                         \\
CUB-10  & 1.000   & 1.000                                                                               & 1.000                                                                                   & 0.289                                                         \\
CUB-15  & 1.000   & 1.000                                                                               & 1.000                                                                                   & 0.168                                                         \\
CUB-20  & 1.000   & 1.000                                                                               & 1.000                                                                                   & 0.114                                                         \\
CUB-25  & 1.000   & 1.000                                                                               & 0.933                                                                                   & 0.092                                                         \\
CUB-50  & 1.000   & 1.000                                                                               & 0.636                                                                                   & 0.044                                                         \\
CUB-100 & 1.000   & 1.000                                                                               & 0.632                                                                                   & 0.020                                                         \\
CUB-200 & N/A     & 1.000                                                                               & 0.225                                                                                   & 0.010                                                        
\end{tabular}
\caption{Average test Jaccard index score comparison across different CUB subsets/dataset, for FastLAS, trained neural DNF-EO model without constraint layer C2 (no post-training processing), the rules extracted from post-training processed neural DNF-EO model without layer C2, and an MLP if interpreted true and false with a threshold of 0. As we have discussed above, MLP with cross-entropy does not provide any mutual exclusivity guarantee in its output and it is expected to see MLP fail to perform when we try to interpret it under a threshold. FastLAS runs out of memory for CUB-200.}
\label{tab:cub-jacc-comp}
\end{table}

\begin{figure}[h]
    \centering
    \includegraphics[width=0.9\textwidth]{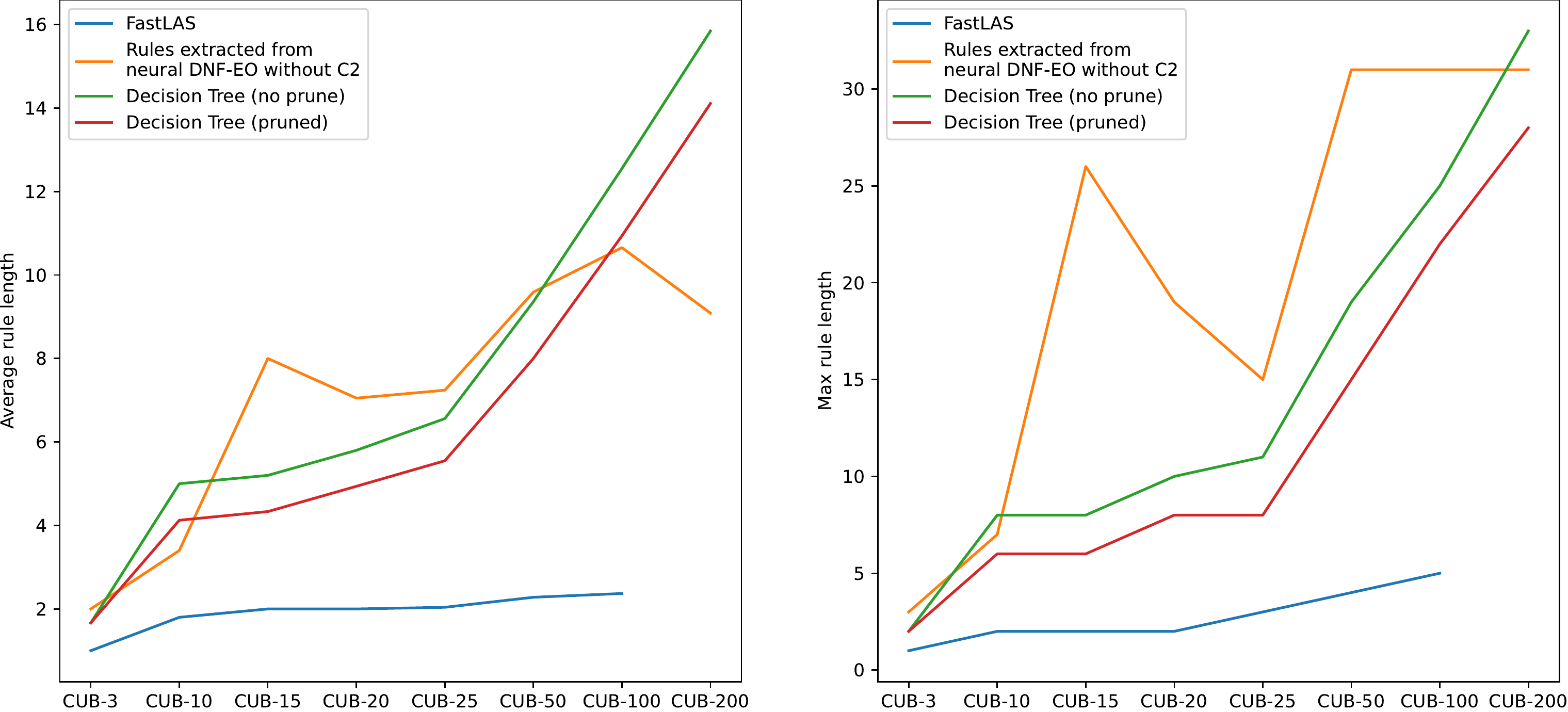}
    \caption{Average rule length and maximum rule length comparison across rules learned from FastLAS, extracted rules from trained neural DNF-EO model and decision tree (both pruned and without pruning). The pruned decision trees still maintain an accuracy of 80\% in train sets. FastLAS runs out of memory for CUB-200.}
    \label{fig:cub-rules-comp}
\end{figure}

\textbf{Interpretability comparison}\ \ \ \ We use the average and maximum rule length as a measurement of interpretability. To measure the rule length of a decision tree, we treat each decision path to a leaf node as the body of a rule, and the number of steps for a decision path can be seen as the number of body atoms. Figure~\ref{fig:cub-rules-comp} shows the comparison of average and maximum rule length across FastLAS' rules, rules extracted from neural DNF-EO model (without layer C2), decision trees without pruning, and pruned decision trees with 80\% accuracy on train sets. FastLAS learns the most compact rules but has scalability issues with a large hypothesis search space, as it fails to learn with 200 classes present. While the maximum rule length for rules extracted from the plain neural DNF is the highest in most experiments, the average length increases linearly compared to decision trees' exponential growth. Rules from our neural DNF-EO are more compact than the decision tree in half of the settings, especially in experiments with large numbers of classes.

\FloatBarrier

\subsection{TMC dataset/subsets} \label{sec:tmc-experiments}

We use macro F1 as the metric for the performance of vanilla neural DNF models and the extracted rules. We train the vanilla neural DNF model with binary cross-entropy across different TMC subsets/dataset, and the classification performance is shown in Table~\ref{tab:dnf-tmc}. The results suggest that our vanilla neural DNF model struggles to learn when more labels are present in the subsets/dataset. However, we speculate that the lack of general patterns/regularities in the TMC dataset/subsets could be the reason for its poor performance. We further discuss this issue in the following section where we compare our model to other baselines. On the other hand, we observe that the rules extracted through the post-training process show very similar performance as the corresponding trained vanilla neural DNF model. Compared to CUB experiments, the thresholding process does not need to maintain mutual exclusion across different outputs and thus does not cause a significant performance drop when `discretising' the continuous weights.

\begin{table}[h]
\begin{tabular}{c|cccccc}
Dataset & \begin{tabular}[c]{@{}c@{}}Train\\ macro F1\end{tabular} & \begin{tabular}[c]{@{}c@{}}Prune\\ macro F1\end{tabular} & \begin{tabular}[c]{@{}c@{}}Finetune\\ macro F1\end{tabular} & \begin{tabular}[c]{@{}c@{}}Threshold\\ macro F1\end{tabular} & \begin{tabular}[c]{@{}c@{}}Rules\\ macro F1\end{tabular} \\ \hline
TMC-3   & 0.745                                                    & 0.764                                                    & 0.739                                                       & 0.754                                                        & 0.754                                                    \\
TMC-5   & 0.615                                                    & 0.649                                                    & 0.648                                                       & 0.635                                                        & 0.635                                                    \\
TMC-10  & 0.391                                                    & 0.380                                                    & 0.370                                                       & 0.362                                                        & 0.362                                                    \\
TMC-15  & 0.177                                                    & 0.143                                                    & 0.100                                                       & 0.097                                                        & 0.097                                                    \\
TMC-22  & 0.094                                                    & 0.086                                                        & 0.093                                                           & 0.094                                                            & 0.094                                                       
\end{tabular}
\caption{\textbf{Vanilla neural DNF} trained with \textbf{binary cross entropy} in different TMC subsets/dataset, at different stages of the training pipeline. All metrics are computed on the test set.}
\label{tab:dnf-tmc}
\end{table}

\noindent \textbf{Comparison to baselines} \label{sec:tmc-experiments-baseline-comparison}\ \ \ \ We report the test macro F1 score of the 2-layer MLP trained with binary cross-entropy loss and the decision tree (without limiting depth) in Table~\ref{tab:mlp-dt-tmc}. There is an increasing performance gap between our vanilla neural DNF model and the MLP as the number of labels increases. Figure~\ref{fig:tmc-dnf-loss} shows the change in macro F1 and $\delta$ magnitude in the vanilla neural DNF model's symbolic layers over 300 epochs of training on TMC-22, and we see that the model achieves a similar macro F1 (around 0.45 compared to MLP's 0.545) when the logical biases are not heavily enforced. But as the $\delta$ magnitude increases, especially over 0.5, the performance of the model drops and never recovers. This suggests that the logical biases in the semy-symbolic layers make it more difficult for the vanilla neural DNF model to learn.

\begin{table}[h]
\begin{minipage}{.43\linewidth}
\begin{tabular}{c|cc}
Dataset & \begin{tabular}[c]{@{}c@{}}MLP + BCE\\ Macro F1\end{tabular} & \begin{tabular}[c]{@{}c@{}}Decision Tree\\ (no pruning)\\ Macro F1\end{tabular} \\ \hline
TMC-3   & 0.817                                                        & 0.925                                                                           \\
TMC-5   & 0.746                                                        & 0.894                                                                           \\
TMC-10  & 0.539                                                        & 0.825                                                                           \\
TMC-15  & 0.538                                                        & 0.863                                                                           \\
TMC-22  & 0.545                                                        & 0.865                                                                          
\end{tabular}
\end{minipage}%
\begin{minipage}{.57\linewidth}
\includegraphics[width=\linewidth]{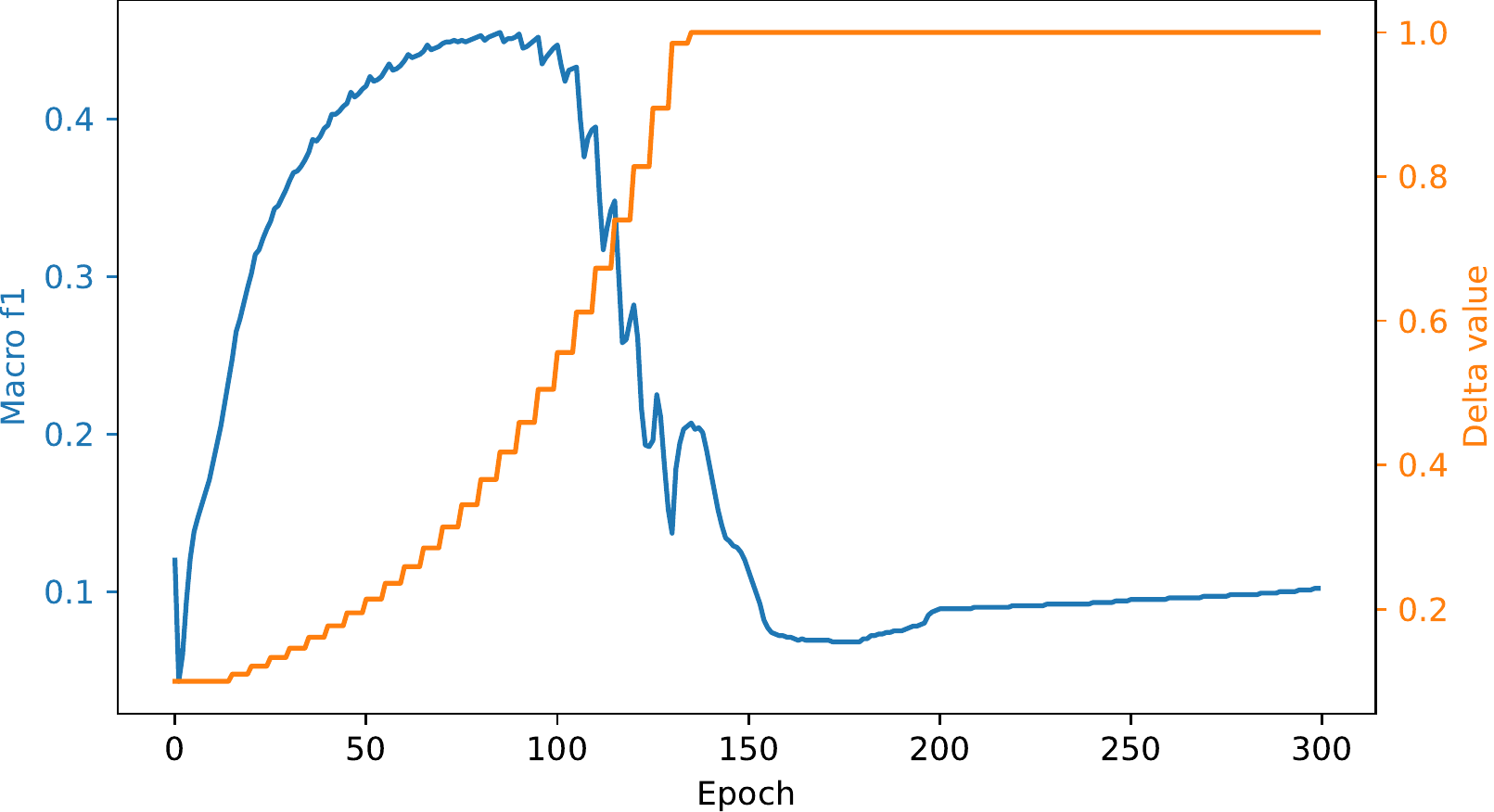}
\end{minipage}
\par
\begin{minipage}[t]{.43\linewidth}
\captionof{table}{Macro F1 score of MLP trained with\\binary cross entropy and decision tree\\without limiting depth on TMC test sets.}
\label{tab:mlp-dt-tmc}
\end{minipage}%
\begin{minipage}[t]{.57\linewidth}
\captionof{figure}{Macro F1 and $\delta$ value of vanilla neural DNF on TMC-22\\over 300 epochs of training. The drop in macro F1 occurs\\around $\delta = 0.5$.}
\label{fig:tmc-dnf-loss}
\end{minipage}
\end{table}

On the other hand, the decision tree learns significantly better than both our vanilla neural DNF model and MLP in all subsets/dataset, as shown in Table~\ref{tab:mlp-dt-tmc}. This suggests that there are very few general patterns to learn in this dataset and that the decision is overfitting to the data. For the neural DNF-based model, if we allow an infinite number of conjunctions (i.e. infinite width of the conjunctive layer), we can encode every possible rule body for a label to cover all cases. But if we do so, the model will overfit: it will learn individual rules that are only `fired' for a small number of data points. The same applies to MLP: an MLP with a wide hidden layer approximates a more complex function that might overfit than an MLP with a small hidden layer. By limiting the number of conjunctions/the width of the hidden layer, the model learns to generalise among large quantities of samples. If the data have very few general patterns, meaning they cannot be encoded using features with low dimensionality, the model will need a wider hidden layer to increase the feature space's dimension and the chance of overfitting increases. The decision tree splits based on input space, and if not limiting the maximum depth of the tree, it can encode every case with nested if-else clauses and overfits the data. To see the difference in model size, we compare the number of trainable parameters of the MLP/vanilla neural DNF model and the number of non-leaf nodes (split nodes) in a decision without limiting depth in Table~\ref{tab:tmc-param-comp}. The decision tree without depth limit has more split nodes (parameters) than its corresponding MLP/vanilla neural DNF in all the experiments. And when we limit the decision tree's max depth to 15 in TMC-22, the tree has 2810 non-leaf nodes and the test macro F1 drops to 0.565 which is close to MLP's performance, confirming that the decision tree without a limit on depth is overfitting. This also shows that the dataset is too complex to learn without increasing the parameters.

The last baseline approach, FastLAS, fails in \textbf{every} TMC subsets/dataset, suffering from the same memory issue as the CUB-200 experiment. Without a mutual exclusivity constraint, the hypothesis search space is exponentially larger since every label can be fired at the same time, causing the memory issue for FastLAS.

\begin{table}[h]
\begin{tabular}{c|cc|cc}
Dataset & \begin{tabular}[c]{@{}c@{}}MLP/Vanilla Neural DNF\\ Architecture\\ (Input, Hidden, Output)\end{tabular} & \begin{tabular}[c]{@{}c@{}}Num. Parameters\\ (Excluding bias)\end{tabular} & \begin{tabular}[c]{@{}l@{}}Decision Tree\\ Max. depth\end{tabular} & No. Non-leaf Nodes \\ \hline
TMC-3   & (59, 9, 3)                                                                                              & 558                                                                        & 54                                                                 & 1461               \\
TMC-5   & (60, 15, 5)                                                                                             & 975                                                                        & 50                                                                 & 3436               \\
TMC-10  & (34, 30, 10)                                                                                            & 1320                                                                       & 34                                                                 & 7715               \\
TMC-15  & (80, 34, 15)                                                                                            & 4275                                                                       & 67                                                                 & 9679               \\
TMC-22  & (103, 66, 22)                                                                                           & 8250                                                                       & 66                                                                 & 11878             
\end{tabular}
\caption{MLP and vanilla neural DNF model's trainable parameters count, comparing to the number of non-leaf nodes in a decision tree without limiting depth, in different TMC subsets/dataset. Non-leaf nodes of a decision tree can be treated as parameters of the decision tree since it encodes a split condition that directs the decision path.}
\label{tab:tmc-param-comp}
\end{table}

\FloatBarrier

\section{Analysis} \label{chap:analysis}

We analyse our neural DNF-based models further in the three areas of \textit{classification performance}, \textit{interpretability}, and \textit{scalability} to compare against the baselines of MLPs, decision trees and FastLAS \cite{fastlas}.

\textbf{Classification performance}\ \ \ \ Our neural DNF-based models learns as well as their MLP counterparts when general patterns exist in the dataset\footnote{To further support this statement, we also construct synthetic datasets for both multi-class and multi-label classification tasks. The data points are generated based on a set of rules such that the data points follow general patterns/show regularities. We train neural DNF-based models on them, and in both scenarios, our models learn almost perfectly, and we can extract symbolic rules. See more details in Appendix~\ref{app:synthetic}}. In CUB experiments, our neural DNF-EO model always achieves test accuracy of 1 as the MLP (Table~\ref{tab:cub-acc-comp}), but also gives test Jaccard index score of 1 and provides mutual exclusion which an MLP trained with cross-entropy cannot provide. Although the TMC experiments show a difference in performance, we speculate that the dataset lacks general patterns/regularities as discussed in Chapter~\ref{sec:tmc-experiments-baseline-comparison}, and the loss plot (Figure~\ref{fig:tmc-dnf-loss}) suggests that the logically constrained bias in our vanilla neural DNF model increases the difficulty of training. In regards to decision trees, our models have a different learning mechanism, and TMC dataset/subsets do not fully reflect the end-to-end learning capability of our models: the data instances in the dataset/subsets cannot be easily encoded in features with low dimensionality, making it hard for our models (and also MLPs) to generalise without overfitting, while decision trees without a depth limit overfit. But when data points in the dataset/subset follow some general patterns, such as the CUB dataset/subsets, our models perform equally well as decision trees (Table~\ref{tab:cub-acc-comp}).

\textbf{Interpretability}\ \ \ \ The logically constrained bias in the semi-symbolic layers makes it possible to extract symbolic rules from our models through post-training processes, contrasting to the lack of interpretability in MLPs. However, as shown in the CUB experiments (Table~\ref{tab:dnf-eo-cub}), our current process does not always result in a successful rule extraction for multi-class classification. The thresholding process is the current bottleneck of the pipeline, as it cannot maintain mutual exclusivity across a large number of classes. Our rules are also not guaranteed to be in the most compact form compared to FastLAS (Figure~\ref{fig:cub-rules-comp}). FastLAS with a scoring function on body length will guarantee the most compact hypothesis, while we do not have a strong signal for the neural DNF-based models to learn the most compact rules, which may further restrict gradient-based learning. Although we do include an L1 regulariser during training and prune the model at post-training processes, it does not necessarily reject hypotheses with longer rules. On the other hand, decision trees show less compactness than rules extracted from our neural DNF-based models, and also are prone to overfitting. 

\textbf{Scalability}\ \ \ \ Our neural DNF-based models scale better than FastLAS. The conjunctive and disjunctive layers of the neural DNF-based model create a hypothesis space that allows rules with any combination of body atoms, and through gradient descent, the model moves towards an approximated hypothesis. FastLAS first needs to compute a hypothesis sub-space that guarantees to contain the optimal solution and may run out of memory during this computation process, which is what happens in the CUB-200 experiment and all TMC experiments. Comparing our model to MLP and decision tree, all three are trainable under the same setting regardless of the input/output size. However, training the neural DNF-based model is more difficult than an MLP (Figure~\ref{fig:tmc-dnf-loss}), and the thresholding process in the post-training pipeline does not yield perfect rules from the trained model anymore after 20 classes (Table~\ref{tab:dnf-eo-cub}).

To summarise, (i) our neural DNF-based model performs as well as an MLP and a decision tree when general patterns exist in the dataset; (ii) it has better interpretability than an MLP by providing ASP rules, and the rules are in average more compact than a decision tree's decision paths; (iii) it also scales better than FastLAS when the hypothesis search space is large.

\section{Related Work}

Our work relates to the research of differentiable Inductive Logic Programming (differentiable ILP). Differentiable ILP methods approximate discrete logical connectives and input with continuous ones and enable continuous logic reasoning, so that the model can be trained via gradient descent. In this category, we can further split the methods into two branches: rule-template-based differentiable ILP and free-form differentiable ILP.

Rule-template-based differentiable ILP methods rely on a set of rule templates (or meta-rules). The rule templates specify the form of rules to learn, and they help with scaling by restricting the hypothesis search space. They also guide the symbolic interpretation of the model. $\delta$-ILP \cite{dnl-ilp}, Neuro-symbolic ILP with Logical Neural Networks (NS-ILP-LNN) \cite{lnn-ilp} and Hierarchical Rule Induction (HRI) \cite{hri} are examples that use rule templates as part of their model instantiation. $\delta$-ILP utilises product t-norm for fuzzy conjunction for differentiable forward chaining, but its rule templates only allow exactly 2 atoms in a rule body. NS-ILP-LNN uses Logical Neural Network \cite{lnn} to constrain the neuron weights and achieve logical behaviour, and the model is based on a tree-structured template. HRI introduces \textit{proton-rule} which extends normal meta-rule and allows flexible predicate arity, making the template set less task-specific. Its hierarchical model follows an embedding-similarity-based inference technique based on \cite{lri}. The choice of rule templates in these approaches requires careful human engineering to ensure a solution would be included in the hypothesis space, and thus the meta-rules/proton-rules set is always task-specific to some extent. Our neural DNF-based model does not rely on rule templates, nor does it restrict the form of rule body: we allow any number of atoms in the rule body and the input predicates of our conjunctive semi-symbolic layer can have any arity as long as grounded\footnote{Although all our experiments are propositional, pix2rule's experiments show the neural DNF module learning from nullary, unary and binary predicates. And to handle first-order logic, we only need to permutate all grounding for each predicate and use grounded predicates as input.}. The only hyperparameter and restriction in our model is the number of conjunctions that can be used by a disjunction.

Free-form differentiable ILP methods do not restrict the form of rules to learn, avoiding the need for human-engineered templates. Neural Logic Machine (NLM) \cite{nlm} is a multi-layer multi-group architecture that realises forward chaining. It is able to learn from nullary, unary and binary predicates, but the learned rules cannot be extracted into a symbolic form. Differentiable Logic Machine (DLM) \cite{dlm} builds upon NLM's architecture but replaces MLPs with logic modules which implement fuzzy logic, and provides symbolic rule extraction after training. Negation is separately applied on conjunction/disjunction outputs in DLM, and it requires tricks such as dropout and Gumbel noise to be trained well. The neural DNF module proposed in pix2rule \cite{pix2rule} handles the negation within a layer without special treatment, and symbolic rules in ASP encoding can be extracted after training. However, pix2rule only demonstrates that the neural DNF module performs well in synthetic binary classification datasets. Our work addresses the limitation by building upon its neural DNF module and experimenting in complex real-world multi-class and multi-label classification tasks, and our models still achieve good performance and provide interpretability by enabling rule extraction.

Another design pattern in the field of neuro-symbolic rule learning is to use separate neural and logical models, which generally does not allow full back-propagation. Two recent works in this category are $Meta_{Abd}$ \cite{metaabd} and NSIL \cite{nlm}. Both methods train a neural network for handling raw data input while their symbolic learners induce logic rules at the same time. $Meta_{Abd}$ adds an abduction procedure and learns with Metagol learning system \cite{metagol}, which is limited to definite logic (no function/constraint etc.). NSIL uses FastLAS \cite{fastlas} to learn from ASP programs with better expressiveness, as well as NeurASP \cite{neurasp} to train the network..

\section{Conclusion and Future Work}

Building upon pix2rule \cite{pix2rule}, we propose a novel neural DNF-EO model to support learning and symbolic rule extraction in real-world multi-class classification, while using the vanilla neural DNF model similar to pix2rule's neural DNF module for multi-label classification tasks. Our evaluations show that when the data have underlying general patterns, our neural DNF-based models learn as well as the MLPs while also providing symbolic rules after post-training processes. Our neural DNF-EO model also provides mutual exclusivity, which MLPs with cross-entropy loss cannot provide. Our models also scale better than FastLAS \cite{fastlas} over large hypothesis search space and can be trained with large numbers of inputs/outputs. Compared to decision trees, our models learn more compact rules and provide end-to-end training capability with other differentiable models.

In future work, we aim towards end-to-end neuro-symbolic rule learning with our neural DNF-based models from unstructured input, such as the task of learning from images in the CUB dataset \cite{cub-200-201}. By connecting our model to other neural networks, the full architecture will first translate low-level signals to high-level representation and then learn symbolic rules. Apart from this, the neuro DNF-EO model's perfect Jaccard index score in the multi-class CUB-20 experiment (48 attributes and 20 classes) gives a promising direction of taking the model into reinforcement learning environments for interpretable policy learning. Many reinforcement environments, such as maze navigation and driving simulation \textit{highway-env} \cite{highway-env}, provide a small number of action choices and the agents can only pick exactly one action at a time. The neural DNF-EO model's symbolic enforcement of mutual exclusivity and its interpretability would make it more favoured than vanilla deep learning approaches. In another direction, taking our neural DNF-based models back to ILP benchmarks would give a better picture of our models' learning capability compared to other classical ILP and differentiable ILP approaches. 

\clearpage

% \begin{acknowledgments}
%   Thanks to the developers of ACM consolidated LaTeX styles
%   \url{https://github.com/borisveytsman/acmart} and to the developers
%   of Elsevier updated \LaTeX{} templates
%   \url{https://www.ctan.org/tex-archive/macros/latex/contrib/els-cas-templates}.  
% \end{acknowledgments}

%%
%% Define the bibliography file to be used
\clearpage
\bibliography{sample-ceur}

\clearpage
\appendix

\section{Data Pre-processing}

\subsection{Procedure on CUB Dataset/subsets} \label{app:data-pre-processing-cub}

Algorithm~\ref{algo:data-pre-processing} shows the pseudo-code of the data pre-processing procedure used by us and also in the Concept Bottleneck Models paper \cite{cbm}. After such pre-processing, all samples of the same class would have the exact same attribute encoding. In the Concept Bottleneck Models paper, the encoding for each class is computed from the training set and also used as the encoding for the validation set and test set as well. Although it is an odd decision to change the validation set and test set using results from the training set, the data pre-processing does ensure attribute label consistency both within each class and across samples and in all 3 sets. We follow this decision in our pre-processing for a fair comparison. The $N$ values we choose for CUB-3, -10, -15, -20, -25, -50, -100, and -200 are 1, 2, 3, 3, 3, 5, 7, and 10 respectively.

\begin{algorithm}[h]
\SetKwInput{Output}{Hyperparameters}
\KwData{\\
\ \ Dataset $D$\\
\ \ Total number of classes $n_c$\\
\ \ Total number of attributes in CUB $n_a$\\
}
\SetKw{Continue}{continue}
\Output{\\
\ \ $N$ -- number of classes an attribute should be present consistently}
\BlankLine

\tcc{Count for each attribute in each class how many times it is present/not present}
$class\_attribute\_count \leftarrow n_c \times n_a \times 2 $ array of all zeros\;
\For{$d \in D$}{
    \For{Attribute $a \in d.attributes$}{
        \If{$a.is\_present == 0 \land a.certainty == 1$}{
            \tcc{Ignore attribute that is not present because not visible}
            continue\;
        }
        $class\_attribute\_count[d.class][a.attribute\_id][a.is\_present] += 1$\;
    }
}
\BlankLine
\tcc{For each class, for each attribute, which presence label (present or not present) is less frequent}
$class\_attribute\_min\_presence \leftarrow \arg\min(class\_attribute\_count, axis=2)$\;
\tcc{For each class, for each attribute, which presence label (present or not present) is more frequent}
$class\_attribute\_max\_presence \leftarrow \arg\max(class\_attribute\_count, axis=2)$\;
\BlankLine
\tcc{Find where most and least frequent, set most frequent to 1}
$equal\_count\_index \leftarrow $ indices where $class\_attribute\_min\_presence == class\_attribute\_max\_presence$\;
$class\_attribute\_max\_presence[equal\_count\_index] \leftarrow 1$\;
\BlankLine
\tcc{Count number of times each attribute is mostly present for a class}
$attribute\_class\_count \leftarrow \text{sum}(class\_attribute\_max\_presence, axis=0)$\;
\BlankLine
\tcc{Select the attributes that are present most of the time on a class level, in at least $N$ classes}
$mask \leftarrow $ indices where $attribute\_class\_count \geq N$
\BlankLine
\tcc{Get the attribute median}
$class\_attribute\_median \leftarrow class\_attribute\_max\_presence[:, mask]$\;
\caption{Data pre-processing for CUB dataset}
\label{algo:data-pre-processing}
\end{algorithm}

\subsection{Procedure on TMC Dataset/subsets} \label{app:data-pre-processing-tmc}

\newcommand{\mi}[2]{\mathrm{I}(#1;#2)}
\newcommand{\entropy}[1]{\mathrm{H}(#1)}
\newcommand{\condEntropy}[2]{\mathrm{H}(#1|#2)}

We calculate the mutual information (MI) of each attribute with respect to the output label combinations, and filter out attributes with MI value less than a threshold $t$. The $t$ values we choose for TMC-3, -5, -10, -15, and -22 are 0.01, 0.02, 0.04, 0.04 and 0.05 respectively. Below we provide the mathematical formulae to calculate the mutual information.

Mutual information of two discrete random variables $X$ and $Y$ can be defined as:

\begin{equation} \label{eq:mi}
    \mi{X}{Y} \equiv \entropy{X} - \condEntropy{X}{Y}
\end{equation}

where $\entropy{X}$ and $\entropy{Y}$ are entropies, and $\condEntropy{X}{Y}$ is conditional entropy. $\entropy{X}$ is calculated as:

\begin{equation} \label{eq:hx}
    \entropy{X} = - \sum_{x \in \mathcal{X}} p(x) \log p(x)
\end{equation}

And the conditional entropy $\condEntropy{X}{Y}$ is calculated as:

\begin{equation} \label{eq:hxy}
    \condEntropy{X}{Y} = - \sum_{y \in \mathcal{Y}} p(y) \sum_{x \in \mathcal{X}} p(x|y) \log p(x|y)
\end{equation}

For each attribute $X_i$, the mutual information of it with respect to the output label combination $Y$ is calculated as the following, using Equation \ref{eq:mi},  \ref{eq:hx} and \ref{eq:hxy}:

\begin{equation} \label{eq:mixiy-raw}
    \mi{X_i}{Y} = - \sum_{x_i \in \mathcal{X}_i} p(x_i) \log p(x_i) - \left(- \sum_{y \in \mathcal{Y}} p(y) \sum_{x_i \in \mathcal{X}_i} p(x_i|y) \log p(x_i|y) \right)
\end{equation}

Since each attribute only has two outcomes: present (1) or not present (0), Equation \ref{eq:mixiy-raw} is equivalent to:

\begin{equation}
\begin{split}
    &\mi{X_i}{Y} = - \Bigl(p(X_i = 1) \log p(X_i = 1) + p(X_i = 0) \log p(X_i = 0) \Bigr) \\
    &- \Biggl(- \sum_{y \in \mathcal{Y}} p(y) \Bigl(p(X_i = 1|y) \log p(X_i = 1|y) + p(X_i = 0|y) \log p(X_i = 0|y)\Bigr) \Biggr)
\end{split}
\end{equation}

\clearpage

\section{Experiments}

\subsection{Code} \label{app:code}

We implement the semi-symbolic layer in PyTorch\footnote{\url{https://pytorch.org/}}. The code for our CUB experiments is available at \url{https://github.com/kittykg/neural-dnf-cub}, and the code for our TMC experiments is available at \url{https://github.com/kittykg/neural-dnf-tmc}. Both repositories contain our implementation of the semi-symbolic layer and the neural DNF-based models.

\subsection{Machine Specification}

The machine specifications for our experiments are shown below in Table~\ref{tab:machine-spec}. 

\begin{table}[h]
\begin{tabular}{c|c|c|c}
Model             & CPU                                                                                                                           & GPU                                                                                        & Memory                 \\ \hline
Neural DNF-based models & \multirow{3}{*}{\begin{tabular}[c]{@{}c@{}}Intel\textregistered\ Xeon\textregistered\ CPU\\ E5-1620 v2 @3.70Ghz\end{tabular}} & \multirow{2}{*}{\begin{tabular}[c]{@{}c@{}}Nvidia GeForce GTX\\ (8GB Memory)\end{tabular}} & \multirow{4}{*}{16 GB} \\
MLP               &                                                                                                                               &                                                                                            &                        \\ \cline{1-1} \cline{3-3}
Decision Tree     &                                                                                                                               & -                                                                                          &                        \\ \cline{1-3}
FastLAS           & \begin{tabular}[c]{@{}c@{}}Intel\textregistered\ Core\texttrademark\\\ i7-7700K CPU @ 4.20GHz\end{tabular}                    & -                                                                                          &                       
\end{tabular}
\caption{Machine specifications for our experiments.}
\label{tab:machine-spec}
\end{table}

\subsection{CUB Experiments}

\subsubsection{Hyperparameters and Model Architecture}

All CUB experiments are run under a random seed of 73. We use Adam \cite{adam-opt} as the optimiser with a learning rate of 0.001 and weight decay of 0.00004. The schedule steps in every 100 epochs. The model architectures, training batch size and training epoch of our neural DNF-EO models are shown in Table~\ref{tab:cub-hyperparameters}. The untrainable constraint layer C2 is of $NUM\_CLASSES \times NUM\_CLASSES$, where relevant connections are set to -6 and rest as 0. We set the pruning threshold $\epsilon=0.005$ in the post-training pipeline.

\begin{table}[h]
\begin{tabular}{c|ccc}
Dataset & In x Conj x Out & Batch size & Epoch \\ \hline
CUB-3   & 34 * 9 * 3      & 4          & 100   \\
CUB-10  & 41 * 30 * 10    & 16         & 100   \\
CUB-15  & 40 * 45 * 15    & 32         & 100   \\
CUB-20  & 48 * 60 * 20    & 32         & 100   \\
CUB-25  & 50 * 75 * 25    & 32         & 100   \\
CUB-50  & 61 * 150 * 50   & 32         & 100   \\
CUB-100 & 82 * 300 * 100  & 32         & 150   \\
CUB-200 & 112 * 600 * 200 & 32         & 200  
\end{tabular}
\caption{Neural DNF-EO model's architecture, training batch size and training epoch for CUB experiments.}
\label{tab:cub-hyperparameters}
\end{table}

The MLP model in each experiment has the same architectural design and training settings as the neural DNF-EO model, but each layer is a normal linear feed-forward layer instead of the semi-symbolic layer.

\subsubsection{Vanilla Neural DNF Models in Multi-class Classification} \label{app:vndnf-cub}

We have also carried out experiments with the vanilla neural DNF model on CUB-3, CUB-10 and CUB-200, and the results are shown in Table~\ref{tab:vanilla-dnf-cub}. With binary cross entropy as a loss function on only CUB-3, the vanilla neural DNF model is able to learn and give a set of rules which results in an average of 1 Jaccard index score on the test set. This is expected since there are only 3 classes to predict (no need for a shortcut by predicting everything as false) and each class is pushed to be binary (true or false). Yet binary cross entropy as a loss function does not work when more classes are present, since it is easier to `cheat' by predicting every class to be false. Cross-entropy loss emphasises the difference between output but does not push the output value to be mutually exclusive, as discussed in Chapter~\ref{chap:cross-entropy-and-multi-class-classification}. Without any enforcement on exactly one class being true, the vanilla neural DNF model with cross-entropy loss is able to predict the correct class under picking the max output (average of 1 accuracy on the test set), but cannot give an output where only one class is close to 1 and everything else is -1 (less than 1 average Jaccard score). These results show the importance of the constraint layer C2 in our neural DNF-EO model.

\begin{table}[h]
\begin{tabular}{c|c|cccccc}
Model + Loss Fn. & Dataset                  & \begin{tabular}[c]{@{}c@{}}Train\\ Acc.\end{tabular} & \begin{tabular}[c]{@{}c@{}}Train\\ Jacc.\end{tabular} & \begin{tabular}[c]{@{}c@{}}Prune\\ Jacc.\end{tabular} & \begin{tabular}[c]{@{}c@{}}Finetune\\ Jacc.\end{tabular} & \begin{tabular}[c]{@{}c@{}}Threshold\\ Jacc.\end{tabular} & \begin{tabular}[c]{@{}c@{}}Rules\\ Jacc.\end{tabular} \\ \hline
Vanilla + BCE    & \multirow{2}{*}{CUB-3}   & 1.000                                                & 1.000                                                 & 1.000                                                 & 1.000                                                    & 1.000                                                     & 1.000                                                 \\
Vanilla + CE     &                          & 1.000                                                & 0.333                                                 & 0.368                                                 & 0.368                                                    & 0.368                                                     & 0.368                                                 \\ \hline
Vanilla + BCE    & \multirow{2}{*}{CUB-10}  & 0.700                                                & 0.417                                                 & 0.433                                                 & 0.433                                                    & 0.433                                                     & 0.433                                                 \\
Vanilla + CE     &                          & 1.000                                                & 0.100                                                 & 0.117                                                 & 0.117                                                    & 0.055                                                     & 0.055                                                 \\ \hline
Vanilla + BCE    & \multirow{2}{*}{CUB-200} & 0.005                                                & 0.001                                                 & -                                                     & -                                                        & -                                                         & -                                                     \\
Vanilla + CE     &                          & 1.000                                                & 0.005                                                 & -                                                     & -                                                        & -                                                         & -                                                    
\end{tabular}
\caption{\textbf{Vanilla neural DNF} trained with \textbf{binary cross entropy} and \textbf{cross entropy} in different CUB subsets/dataset, at different stages of training pipeline. All metrics are calculated by averaging across test set. Vanilla neural DNF model performs poorly when 200 classes are present, and thus we do not proceed with the post-training stage.}
\label{tab:vanilla-dnf-cub}
\end{table}

\subsection{TMC Experiments}

\subsubsection{Hyperparameters and Model Architecture}

All TMC experiments are run under a random seed of 73. We use Adam \cite{adam-opt} as the optimiser with a learning rate of 0.001 and weight decay of 0.00004. The schedule steps in every 100 epochs. Table~\ref{tab:tmc-hyperparameters} shows the model architectures, training batch size and training epoch of our vanilla neural DNF models. We set the pruning threshold $\epsilon=0.005$ in the post-training pipeline.

\begin{table}[h]
\begin{tabular}{c|ccc}
Dataset & In x Conj x Out & Batch size & Epoch \\ \hline
TMC-3   & 59 * 9 * 3      & 4          & 300   \\
TMC-5   & 60 * 15 * 5    & 16          & 300   \\
TMC-10  & 34 * 30 * 10    & 32         & 300   \\
TMC-15  & 80 * 45 * 15    & 32         & 300   \\
TMC-22  & 103 * 66 * 22    & 32        & 300 
\end{tabular}
\caption{Vanilla neural DNF model's architecture, training batch size and training epoch in TMC experiments.}
\label{tab:tmc-hyperparameters}
\end{table}

The MLP in each experiment has the same architectural design as the vanilla neural DNF and training settings, but each layer is a normal feed-forward linear layer instead of the semi-symbolic layer.

\subsection{Synthetic Datasets Experiments} \label{app:synthetic}

We generate synthetic multi-class and multi-label datasets to show that the neural DNF-based models are capable of learning rules from datasets with underlying patterns. For each synthetic dataset, we first randomly generated a set of rules that guarantee mutual exclusivity if under a multi-class setting. Then, 10,000 samples are generated based on the rule set without noise. 2,000 of the samples are randomly selected to form the test set, 1,600 for the validation set and 6,400 for the train set. Following the same training and post-training process but without the tuning stage, we evaluate a neural DNF-based model on each of the synthetic datasets. The result is shown in Table~\ref{tab:synthetic-test} below. We see that our neural DNF-based models are learning well after training in all experiments, while we still observe the problem of multi-class classification performance drop after thresholding.

\begin{table}[h]
\begin{tabular}{c|cccc}
Synthetic Dataset + Model + Metric                                                                 & After train & After prune & After threshold & Extracted Rules \\ \hline
\begin{tabular}[c]{@{}c@{}}Multi-class 3 classes +\\ Neural DNF-EO +\\ Jaccard score\end{tabular}  & 1.000       & 1.000`      & 0.981           & 0.974           \\
\begin{tabular}[c]{@{}c@{}}Multi-class 25 classes +\\ Neural DNF-EO +\\ Jaccard score\end{tabular} & 0.995       & 0.991       & 0.227           & 0.226           \\
\begin{tabular}[c]{@{}c@{}}Multi-label 3 labels +\\ Vanilla Neural DNF +\\ Macro F1\end{tabular}   & 1.000       & 0.995       & 0.995           & 0.995           \\
\begin{tabular}[c]{@{}c@{}}Multi-label 25 labels +\\ Vanilla Neural DNF +\\ Macro F1\end{tabular}  & 1.000       & 0.967       & 0.987           & 0.987          
\end{tabular}
\caption{Neural DNF-based models' performance on different synthetic datasets at different stages of the pipeline. All performance is measured on the test set.}
\label{tab:synthetic-test}
\end{table}

 These synthetic experiments can be found under the \mintinline{bash}{synthetic_test/notebooks/} folder in the two GitHub repositories we mentioned in Appendix~\ref{app:code}.

\end{document}